\documentclass[10pt,twocolumn,letterpaper]{article}

\usepackage{iccv}
\usepackage[american]{babel}
\usepackage{times}
\usepackage{epsfig}
\usepackage{graphicx}
\usepackage{amsmath}
\usepackage{amssymb}
\usepackage{booktabs} 
\usepackage{color} 
\usepackage{setspace}
\usepackage{amsmath,amssymb, amsfonts}
\usepackage[bold]{hhtensor}
\usepackage{multirow} 
\graphicspath{figures/}

\newcommand{\squishlist}{
 \begin{list}{$\bullet$}
  { \setlength{\itemsep}{0pt}
     \setlength{\parsep}{1pt}
     \setlength{\topsep}{1pt}
     \setlength{\partopsep}{0pt}
     \setlength{\leftmargin}{1.5em}
     \setlength{\labelwidth}{1em}
     \setlength{\labelsep}{0.5em} } }
\newcommand{\squishend}{
  \end{list}  }

\DeclareMathOperator*{\argmin}{argmin}

\usepackage[breaklinks=true,colorlinks,bookmarks=false]{hyperref}
\iccvfinalcopy 

\author{Jiacheng Chen\textsuperscript{*1}, Bin-Bin Gao\textsuperscript{*2}, Zongqing Lu\textsuperscript{\dag1}, Jing-Hao Xue\textsuperscript{3}, Chengjie Wang\textsuperscript{2}, Qingmin Liao\textsuperscript{1}\\
{\centerline{\textsuperscript{1}Tsinghua University\qquad
\textsuperscript{2}Tencent Youtu Lab\qquad
\textsuperscript{3}University College London}}\\
\centerline{
{\tt\small cjc19@mails.tsinghua.edu.cn}\quad
{\tt\small gaobb@lamda.nju.edu.cn}\quad
{\tt\small luzq@sz.tsinghua.edu.cn}}\\
\centerline{
{\tt\small jinghao.xue@ucl.ac.uk}\quad
{\tt\small jasoncjwang@tencent.com}\quad
{\tt\small liaoqm@tsinghua.edu.cn}}
}

\def\iccvPaperID{5733} 

\begin{document}

\title{SCNet: Enhancing Few-Shot Semantic Segmentation\\ by Self-Contrastive Background Prototypes}
\maketitle

{
\renewcommand{\thefootnote}{\fnsymbol{footnote}}
\footnotetext[1]{Both authors contributed equally. This work was done when Jiacheng Chen was an intern at Tencent Youtu Lab.} 
\footnotetext[2]{Corresponding author.}
}
\begin{abstract}
Few-shot semantic segmentation aims to segment novel-class objects in a query image with only a few annotated examples in support images. Most of advanced solutions exploit a metric learning framework that performs segmentation through matching each pixel to a learned foreground prototype. However, this framework suffers from biased classification due to incomplete construction of sample pairs with the foreground prototype only. To address this issue, in this paper, we introduce a complementary self-contrastive task into few-shot semantic segmentation.  Our new model is able to associate the pixels in a region with the prototype of this region, no matter they are in the foreground or background. To this end, we generate self-contrastive background prototypes directly from the query image, with which we enable the construction of complete sample pairs and thus a complementary and auxiliary segmentation task to achieve the training of a better segmentation model. Extensive experiments on PASCAL-5$^i$ and COCO-20$^i$ demonstrate clearly the superiority of our proposal. At no expense of inference efficiency, our model achieves state-of-the results in both 1-shot and 5-shot settings for few-shot semantic segmentation.

\end{abstract}

\section{Introduction}\label{sec:intro}
Few-shot semantic segmentation(FSS)~\cite{OSLSM, panet, canet} has attracted many attention because it makes pixel-level semantic predictions for novel classes on testing images~(\emph{query}) with few~(\eg, 1 or 5) labeled images~(\emph{support}). This learning paradigm aims to model quickly adapting ability to novel classes that have never been seen during training and use only a few labeled images on testing. It greatly reduces the requirement for collecting and annotating a large-scale dataset. 

\begin{figure}[t]
\centering
\centerline{\includegraphics[width=1\linewidth]{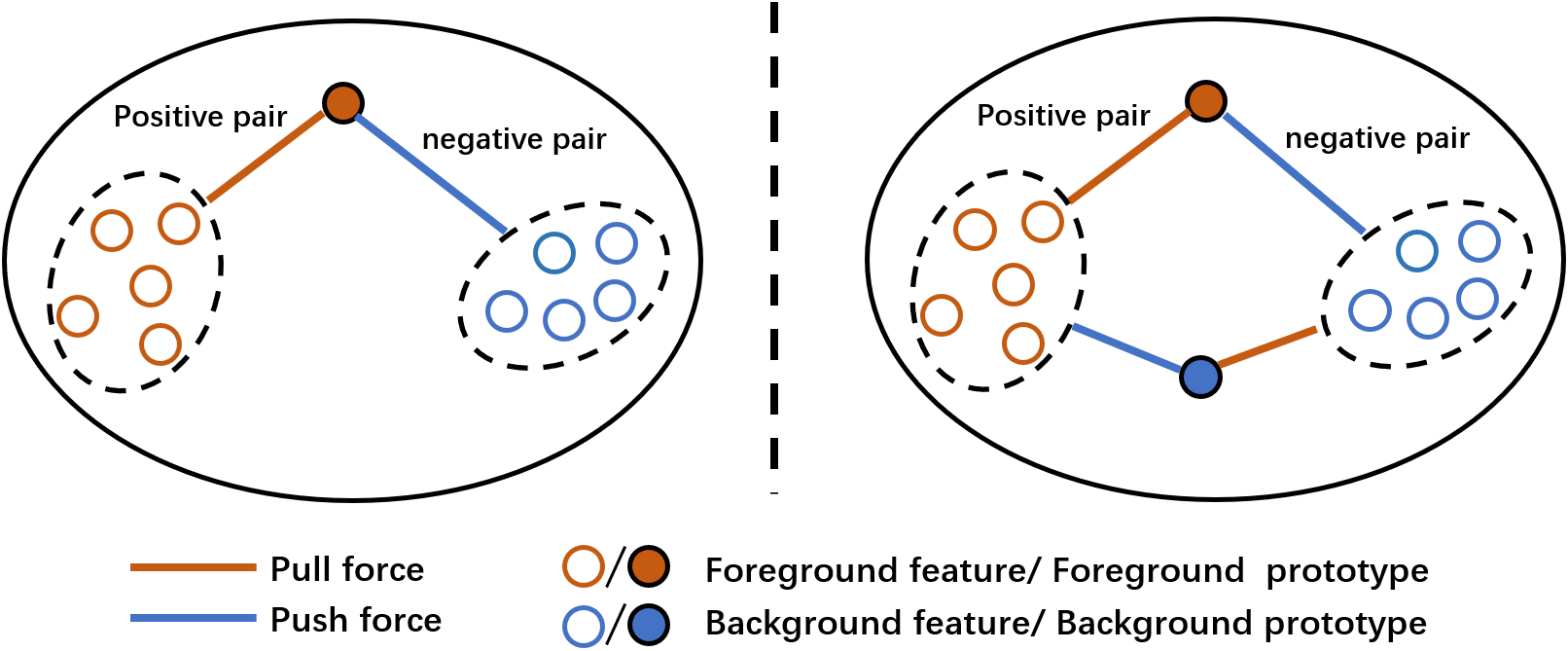}}
\caption{Left: Previous work construct positive pairs and negative pairs for foreground prototypes only. Right: Our SCNet generate background prototypes and is thus able to construct feature pairs for both foreground prototypes and background prototypes.}
\label{fig:point}
\end{figure}

Most FSS methods~\cite{panet, canet, pfenet, fwb, PL} are built on metric learning for its simplicity and effectiveness, by learning to compare query and few support images. First, a shared convolution network is used to concurrently extract deep representations of support and query images. Then, these support features and their masks are encoded to a single vector which forms a class-specific foreground~\emph{prototype}. Finally, pixel-level comparison are densely conducted between the prototype and each location of query features that determines if they are from matching categories or not. This comparison may be explicit,~\eg, cosine-similarity~\cite{PL,panet}, as well as implicit,~\eg, relation network~\cite{sung2018learning}. Following this framework, some works try to generate a more fine prototype with support features~\cite{pmms, PGNet, ppnet}, while PFENet~\cite{pfenet} enriches query features with a multi-scale fusion strategy. However, these methods suffer from limited generalization ability due to incomplete feature comparison. 

\begin{figure}[t]
\centering
\centerline{\includegraphics[width=1.02\linewidth]{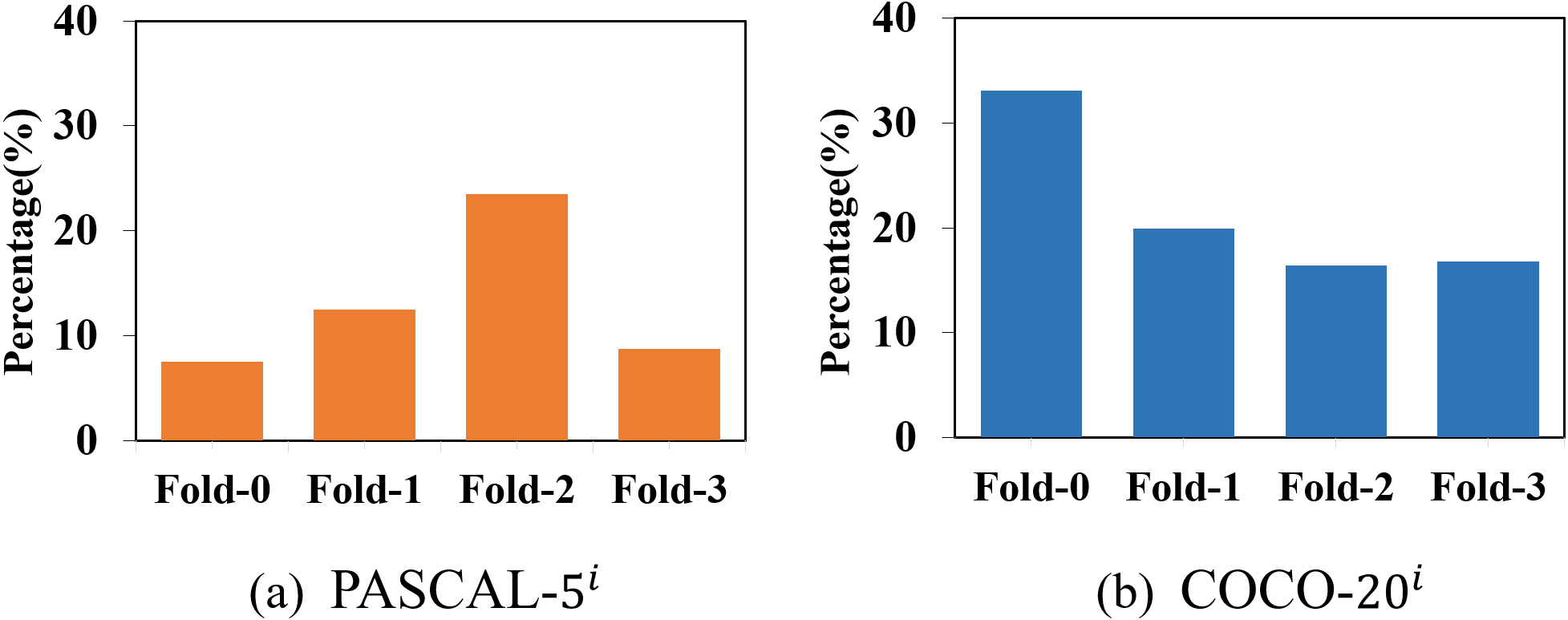}}
\caption{The percentage of novel classes objects in training set (base class). We firstly count the number of basic classes and novel classes on all training images, respectively, and then calculate the percentage of the novel classes for each training fold.}
\label{fig:STASTICS}
\end{figure}

In previous works, only foreground prototypes were generated from the support images. Therefore, as shown in Fig.~\ref{fig:point}, during the training of previous methods, as only (foreground) prototypes are available, the whole background features in the query image are treated as negative samples. 
This will lead to an issue with FSS, because it is possible, and in fact, it is the case, that some novel class objects (in the test set) are present in the base training set and treated as background during training. In Fig.~\ref{fig:STASTICS}, we can see that there is a high percentage of novel-class objects in each base fold. For example, 23.5\% of novel-class objects are hidden in the training images of Fold-2 in PASCAL-5$^i$. Therefore, it is unsurprising that, during the testing, previous methods tend to incorrectly `remember' novel-class objects as background, even when the (foreground) novel-class prototype is provided.

The cornerstone of our solution to this problem is to generate background prototypes in the training episodes and thus construct complete feature pairs for comparison, \ie to additionally construct positive and negative pairs for background prototypes, as shown in Fig.~\ref{fig:point}. In this way, we can mitigate the prior bias that `remembers' novel-class objects in the training set as background. Furthermore, the background is defined as any area other than the annotated objects in FSS. So it is hard to guarantee similar semantics between the backgrounds of support and query images. Unlike previous methods~\cite{simpropnet, panet, pmms} which also generate background prototypes, we propose to extract background prototypes from the query features alone to ensure the similarity between the prototype and the background features during prediction. That is, the comparison among query features is conducted in a self-contrastive manner.

\begin{figure*}[tb]
\centering
\centerline{\includegraphics[width=.98\linewidth]{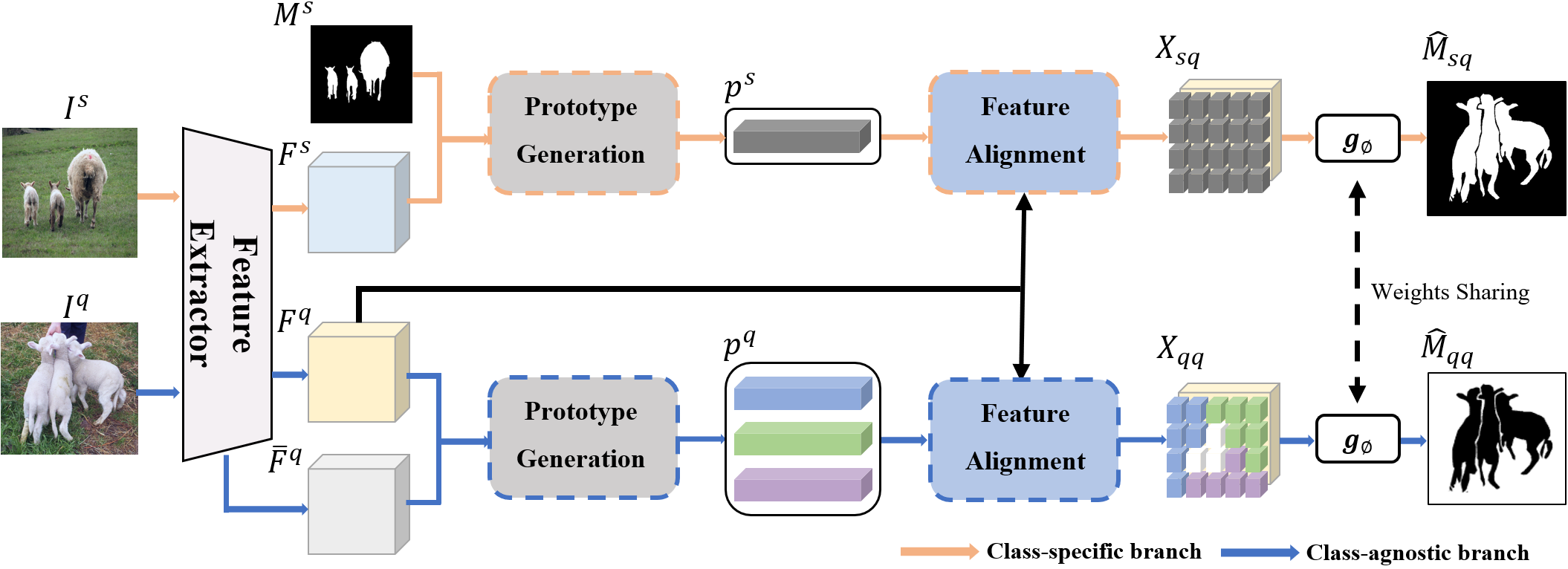}}
\caption{The pipeline of our CSSNet. CSSNet learns complementary features comparison with two parallel branches, \ie, class-specific and class-agnostic branch. The class-specific and class-agnostic prototypes are firstly obtained by their individual prototype generation module. Then, these features comparison is conducted through feature alignment and a shared convolution module. Note that, the goal is approximate but the implementation is very different for prototype generation and feature alignment module.}
\label{fig:overview}
\end{figure*}

Specifically, the pipeline of our proposed SCNet is shown in Fig.~\ref{fig:overview}. 
It consists of two parallel branches, class-specific and class-agnostic branches. Each branch has two sub-modules, prototype generation and feature alignment. Different from other two-branch methods~\cite{simpropnet, crnet}, our two branches learn feature comparison with both foreground and background prototypes. The class-specific branch learns between the foreground prototype from the support and the query features; its goal is to segment the foreground area in the query image. The class-agnostic branch learns among the query features; it encourages the background prototype to pull background features and push away foreground features. To further generate more fine background prototypes, we use a clustering algorithm, $k$-means, to group the entire query feature map and guide feature comparison in a self-contrastive way. The proposed SCNet effectively and efficiently improves the FSS performance without additional parameters and inference costs. 
\squishlist 
\item Firstly, we present a novel SCNet that learns features comparison not only between the prototype of support images and query features but also among query features themselves. To the best of our knowledge, the self-contrastive manner among query features is the first time being proposed to enrich feature comparison and help us to yield an unbiased segmentation model at the few-shot setting.

\item Secondly, we propose a simple but effective self-contrastive learning branch~(class-agnostic) that includes background prototype generation and feature alignment. In practice, this branch can dynamically generate multiple prototypes for the background region of a query image and compare features in a self-contrastive manner.

\item Thirdly, we achieve new state-of-the-art results on both PASCAL-5$^i$ and COCO-20$^i$ datasets without additional parameters and computation cost at inference time. In addition, we also extensively demonstrate the effectiveness of the proposed method.
\squishend
\section{Related Works}

\noindent\textbf{Few-Shot Segmentation}  
Few-shot semantic segmentation aims to perform pixel-level classification for novel classes in a query image conditioned on only a few annotated support images. OSLSM~\cite{OSLSM} first introduces this setting and uses parametric classification to solve this problem. PL~\cite{PL} and PANet~\cite{canet} use prototypes to represent typical information for foreground objects present in the support images, and make predictions by pixel-level feature comparison between the prototypes and query features via cosine similarity. This comparison can also be performed in an implicit way. For example, CANet~\cite{canet} uses convolution to replace the cosine similarity for complex query images.
Recently, many studies try to fully mine support features. PGNet~\cite{PGNet} proposes a graph attention unit that treats each location of the foreground features in the support images as an individual and establishes the pixel-to-pixel correspondence between the query and support features. PMMs~\cite{pmms} uses the prototype mixture model to correlate diverse image regions with multiple prototypes. However, they only consider the prototype extraction from the support features and ignore the prototype representation from the query images. PFENet~\cite{pfenet} exploits multi-scale query features to strengthen its representation ability. 

In this paper, we design a twin-branch deep network, which exploits feature comparison not only between the support and query features but also between the query features themselves and performs two symmetric segmentation tasks to enhance feature comparison.

\noindent\textbf{Self-Supervised Learning}
In recent years, self-supervised learning has made remarkable success in unsupervised representation learning. It aims at designing pretext tasks to generate pseudo labels without additional manual annotations. Typical pretext tasks include predicting transformation parameters~\cite{predictingtransformationparameters1, predictingtransformationparameters2}, contrastive learning~\cite{contrastivelearning1, contrastivelearning2, contrastivelearning3}, and recovering the input under some corruption~\cite{recovering1, recovering2}. Our work is related to clustering-based methods~\cite{clustermethods1, clustermethods2, clustermethods3, clustermethods4}, which use cluster assignments as pseudo-labels to learn deep representations. DeepCluster~\cite{clustermethods3} iteratively groups deep features with a clustering algorithm and uses the subsequent assignments as supervision to update the weights of the network. SeLa~\cite{clustermethods1} takes pseudo-label assignment as an optimal transport problem and conducts simultaneous clustering and representation learning. 

In our work, we generate background prototypes by clustering query features with high-level semantic information, and it is able to supply complementary features with self-contrastive manner.

\section{Our Method}
\subsection{Few-Shot Semantic Segmentation Problem}
A few-shot semantic segmentation (FSS) system to segment the area of unseen class $C_{novel}$ from each query image given few labeled support images. Models are trained on base classes $C_{base}$ (\emph{training set}) and tested on novel classes $C_{novel}$ (\emph{test set}). Notice that $C_{base}$ and $C_{novel}$ are non-overlapping, 
which ensures that the generalization ability of segmentation model to new class can be evaluated.

We adopt the episode training mechanism, which has been demonstrated as an effective approach to few-shot learning. Each episode is composed of a support set $S$ and a query set $Q$ of the same classes. The support set $S$ has $k$ image-mask pairs, \ie, $S=\{(I_{i}^{s},M_{i}^{s} )\}_{i=1}^{k}$, which is termed as ``$k$-shot", with $I_{i}$ the $i$-th support image and $M_i$ its corresponding mask. For query set $Q=\{(I^q,M^q)\}$, where $I^q$ is a query image and $M^q$ is its ground truth mask. The support-query triples $(\{(I_{i}^{s},M_{i}^{s} )\}_{i=1}^{k}, I^q)$ forms the input data of FSS model, and the goal is to maximize the similarity between $M^q$ and the generated prediction $\hat{M}^q$ on $I^q$. Therefore, how to exploit $S$ for the segmentation of $I^q$ is a key task of few-shot segmentation.

To simplify the notation, let us take the ``1-shot" segmentation for example, \ie, $S=(I^{s},M^{s})$. That is, given a triples $(I^{s},M^{s}, I^q)$ input, the goal of our proposed SCNet is to directly learn a conditional probability mass function $\hat {M}^q = p(I^q|(I^{s},M^{s}, I^q); \vec \theta)$ on base set $C_{base}$, where $\vec\theta$ is the parameters in the whole network. We use $(F^s, F^q)$ to denote the feature map of $(I^s, I^q)$ from a CNNs backbone, \eg, VGG-Net or ResNet, where $F^s$ and $F^q \in \mathbf{R}^{h\times w \times c}$.

\subsection{Complementary Features Learning}

As shown in Fig.~\ref{fig:overview},
our SCNet is a two-branch architecture: the class-specific branch learns feature comparison through support-query pairs, while the class-agnostic branch learns feature comparison in the query image itself. 

\subsubsection{Learning to Compare between Support-Query Pair }\label{sec:SQP} 
The foreground prototype $\vec p^s$ 
is generated by mask average pooling (MAP) over the support features at locations $(i,j)$: 
\begin{equation}\label{eq:map}
 \vec p^s = \frac {\sum_{i,j}{\vec F_{i,j}^s \odot M_{i,j}^s}}{\sum_{i,j}{M_{i,j}^s}},
\end{equation}
where $\odot$ is the broadcast element-wised product.
Then, the prototype $\vec p^s$ is assigned to each spatial location on query features $F^q$ and learns feature comparison to identify foreground objects presented in the query image. Technically, we first expand $\vec p^s$ to the same shape as the query features $F^q$ and concatenate them in the channel dimension as
\begin{equation}\label{eq:Xsq}
X_{sq} = \mathcal{C}\big(\mathcal E^s(\vec p^s), F^q\big),
\end{equation}
where $\mathcal{E}$ is the expansion operation, and $\mathcal{C}$ is the concatenation operation between two tensors along the channel dimension.
In order to obtain a better segmentation mask $\hat M^q$ on query image $I^q$, we follow previous methods~\cite{canet, dan}, which usually use a convolution module $g_\phi$ to encode the fusion feature $X_{sq}$, as
\begin{equation}\label{eq:Msq}
\hat M_{sq}^q = g_\phi(X_{sq}).  
\end{equation}
Here, $g_\phi$ performs the pixel-level feature comparison in the way of binary classification, which verifies  whether the pixels in the feature map $F^q$ match the prototypes in $\mathcal{E}^s(\vec{p}^s)$ at the corresponding location. 

\subsubsection{Learning to Compare in Query Image} 
\begin{figure*}[t]
\centering
\centerline{\includegraphics[width=1\linewidth]{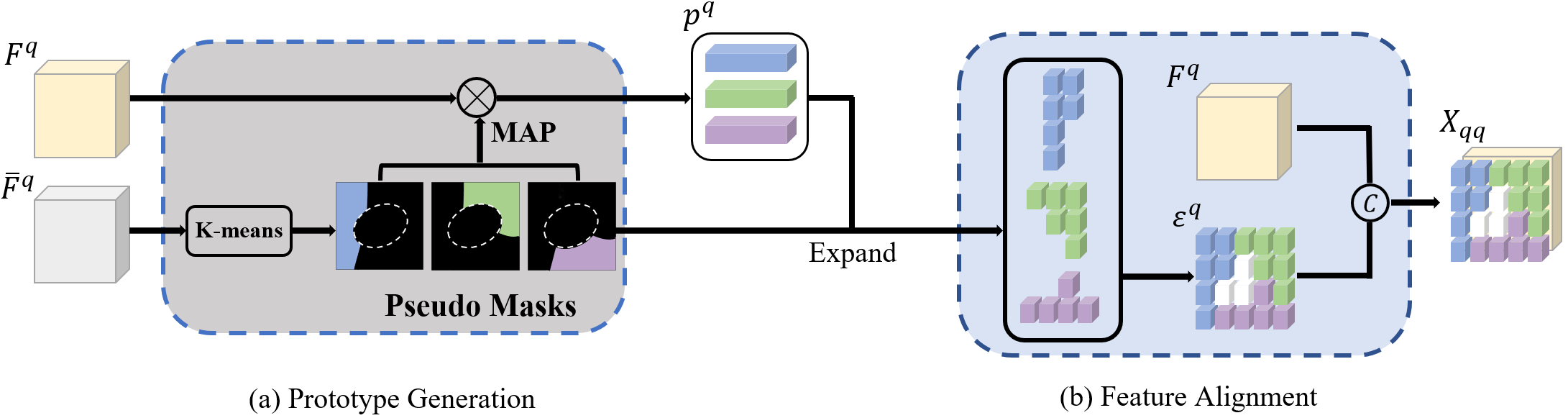}}
\caption{Illustration of our prototype generation and feature alignment module in the class-agnostic branch. The query features $F^q$ is first spatially partitioned into 3 regions by clustering on high-level feature $\bar {F}^q$. Then, the region prototypes $\vec p^q$ are generated by MAP. Finally, the prototypes are expanded to corresponding region, and concatenated with the query features to perform comparison. The white square in $\mathcal E^q$ represent randomly selected prototypes from $\vec p^q$.}
\label{fig:module}
\end{figure*}

Different from the class-specific branch to which support images supply mask annotation for a specific class, our class-agnostic branch learns from query features $F^q$ themselves since the background region of a query image does not consist of any annotation. As diagrammed in Fig.~\ref{fig:module}, to obtain the background prototypes for a query image, we firstly employ $k$-means clustering to group it into $n$ regions (\eg, 3 regions) at the feature level. Then, these region prototypes $\{\vec p_i^q, i=1,2,\cdots,n\}$ are generated by a mask average pooling as in Eq.~(\ref{eq:map}). 
Finally, the convolution module $g_\phi$ is used to compare each feature to the prototypes as discussed in Sec.~\ref{sec:SQP}. 
In this region, the ground truth to supervise $g_\phi$ is a mask of ``1''s that covers the background area, as we concatenate $n$ background prototypes to their corresponding features. 
For area outside the clustering regions, namely the foreground region, assigning it the corresponding foreground prototype will lead to a trivial solution, \ie, a full `1's mask.
We randomly select a background prototype for the foreground region, as shown in the white square in Fig.~\ref{fig:module}(b). Finally, we perform a series of operations, \ie, prototype expansion $\mathcal{E}(\cdot)$, feature concatenation $\mathcal{C}(\cdot)$ and convolution $g_\phi(\cdot)$, as those in the class-specific branch described above. Note that the class-agnostic branch, with its ground-truth mask being $1-M^q$, is learned in a direction opposite the class-specific branch. 

\noindent\textbf{Background Prototypes Generation.} The class-specific branch extracts a prototype with the MAP on the support features and compares it with the query features, because there are same-class objects in both the support and query images. However, it is hard to guarantee that there is similar semantics in the backgrounds of support-query pairs. Thus it is more appropriate to estimate background prototypes directly from the query features. To this end, a naive method is to directly apply global mask average pooling on a background region of the query features to obtain its background prototype. But as the background is usually much more diverse than a particular foreground, so simple mask average pooling is actually ineffective. To alleviate this issue, we propose to spatially partition the query features according to their semantic distribution in the feature space.

As we know, high-level semantic features are usually extracted by the deeper layers of a network. Therefore, instead of using the $F^q$ from the middle layers, we employ higher-level feature $\bar {F}^q$ to group the background of $F^q$. To partition $\bar {F}^q$ into $n$ regions, we use classical $k$-means clustering, the optimization problem of which can be expressed as 
\begin{align}\label{eq:kmeans}
\begin{split}
   & \min_{r_{ik}, \vec u_k} \sum_{i=1}^{hw} \sum_{k=1}^{n} r_{ik} {\rm dist}\left(\vec {\bar{F}}_{i}^{q}, \vec u_{k}\right),\\
&s.t.\ \ \sum_{k=1}^n{r_{ik}}=1, \forall i,
\end{split}
\end{align}
where ${\rm dist}(\cdot)$ is a standard cosine distance, the binary indicator variables $r_{ik}\in\{0,\ 1\}$, $\vec {\bar{F}}_{i}^{q} \in \bar F^q$, and $\vec u_k$ represents the centre of the $k$-th cluster. The optimal solution $r_{ik}^*$
and $\vec u_k^*$ can be readily obtained iteratively.

With $r_{ik}^*$, we can easily obtain the background prototype of the $k$-th cluster region of $F^q$. First, we reshape the $k$-th binary indicator vector $\{r_{ik}\}_{i=1}^{hw}\in \mathbf{R}^{hw}$ to $\bar{M}_k^q\in \mathbf{R}^{h\times w}$. Hence, $\bar{M}_{k}^q$ depicts which spatial locations belong to the $k$-th cluster, and these locations are equal to one while others are zero. Notice that $\bar{M}_{k}^q$ may contain foreground, so we check the intersection of $\bar{M}_k^q$ and the foreground mask ${M}^q$ and update $\bar{M_k}^q$ as    

\begin{equation}
\bar{M_k}^q \gets \bar{M_k}^q-\bar{M_k}^q \cap {M}^q,
\end{equation}
Finally, the $k$-th background prototype is computed by using the MAP operation over locations $(i,j)$:
\begin{equation}
\vec p_{k}^{q}= \frac {\sum_{i, j}{\vec {F_{i, j}^q} \odot  \bar M_{i,j,k}^q}}{\sum_{i, j}{ \bar M_{i,j,k}^q}},
\end{equation}
where $k=1,2,\cdots, n.$

\noindent \textbf{Feature Alignment for Complete Comparison.} 
The class-specific branch only encourages foreground features to be close to each other and far away from the background features of the query image. As discussed above, in the few-shot semantic segmentation setting, this learning mechanism easily forces the model to `remember' objects outside the base class set $C_{base}$ as background during the training phase, which limits the model ability of generalization. Hence in our work, we use a complete feature comparison to mitigate this issue.

Note that now we have obtained $n$ background prototypes from a query image. How to align these prototypes with query features $F^q$? It is natural that a prototype should be as close as possible to its corresponding features and meanwhile as far as possible from the foreground features.

Hence, technically, we first expand each prototype $\vec p_k^q$ to fill the corresponding locations in where $\bar{M}_{i,j,k}^q$ is 1. Then, all the expanded prototypes $\vec p_k^q (k=1,2,\cdots, n)$ and their corresponding query features are concatenated in the depth dimension as Eq.~(\ref{eq:Xqq}), as shown in Fig.~\ref{fig:module}(b). In this way, we achieve the construction of positive pairs for each background prototype. 
In the meantime, to construct negative pairs, we randomly select one from all background prototypes and densely pair it with each location of foreground features in the query image. We denote the above expansion operation for all background prototypes as $\mathcal E^q(\cdot)$, and similarly to Eq.~(\ref{eq:Xsq}) we have
\begin{equation}\label{eq:Xqq}
X_{qq}= \mathcal{C}\big(\mathcal E^q(\vec p_k^q|k=1,2,\cdot,n), F^q\big), 
\end{equation}
Finally, a convolution module $g_\phi$ to encode $X_{qq}$ for learning better comparison metric as
\begin{equation}
\hat M_{qq}^q = g_\phi(X_{qq}).  
\end{equation}
Note that all parameters of $g_\phi$ are shared on both $X_{sq}$ and $X_{qq}$.

\subsection{Learning to Completely Compare}
\noindent\textbf{Loss Function.} 
Note that $\hat M_{sq}^q$ is an encoding output based on $X_{sq}$, which combines foreground prototype with the query features, while $\hat M_{qq}^q$ is obtained by $X_{qq}$, that using the background prototypes and the query features. 
Their common characteristic is the feature comparison between foreground and background, no matter it is $X_{sq}$ or $X_{qq}$. It is a reasonable requirement that if these comparison are on the same semantic (\ie, foreground class or background cluster), the model should output the prediction 1 at these positions and 0 otherwise.
That is, for the class-specific branch, its target label should be the mask $M^q$ of a query image, while for the class-agnostic branch, its target label should be $1-M^q$. 
Hence, we use the cross-entropy loss and formulate the overall loss function as
\begin{equation}
    \mathcal L=(1-\lambda) \mathcal L_{1}\left(\hat{M}_{sq}, M^{q}\right)+\lambda \mathcal L_{2}\left(\hat {M}_{qq}, 1-M^{q}\right)
\end{equation}
where $\lambda$ is a parameter to balance the two cross-entropy losses.
When $\lambda$ is 0, only the $\mathcal L_1$ is left in the overall loss function. Then our SCNet degenerates to the baseline.

\noindent\textbf{Inference.} 
Given a query image and $k$ support images, we take the average of all foreground prototypes from $k$ support images as the new foreground prototype. Note that each branch can yield a segmentation result in the training phase. But during evaluation, the model just outputs $\hat {M}_{sq}$ without $\hat{M_{qq}}$, because the mask of the query image is unknown at the inference stage. Hence no extra inference costs.


\begin{table*}[h]
\caption{Results of 1-shot and 5-shot segmentation on PASCAL-5$^i$ using the mean-IoU. Best results in bold.}
\label{table:pascal}
\resizebox{\textwidth}{!}{
\begin{tabular}{@{}lcccccc|ccccc@{}}
\toprule
                     &          & \multicolumn{5}{c|}{1-Shot}                                                   & \multicolumn{5}{c}{5-Shot}                                                    \\ \cmidrule(l){3-12} 
Methods              & Backbone & Fold-0       & Fold-1       & Fold-2       & Fold-3       & Mean          & Fold-0       & Fold-1       & Fold-2       & Fold-3       & Mean          \\ \midrule
SG-One~\cite{sgone}(TCYB'20)      &          & 40.2          & 58.4          & 48.4          & 38.4          & 46.3          & 41.9          & 58.6          & 48.6          & 39.4          & 47.1          \\
AMP~\cite{amp}(ICCV'19)         &    & 41.9          & 50.2          & 46.7          & 34.7          & 43.4          & 41.8          & 55.5          & 50.3          & 39.9          & 46.9          \\
PANet~\cite{panet}(ICCV'19)       &          & 42.3          & 58.0          & 51.1          & 41.2          & 48.1          & 51.8          & 64.6          & 59.8          & 46.5          & 55.7          \\
RPMM~\cite{pmms}(ECCV'20)        & VGG-16          & 47.1          & 65.8          & 50.6          & 48.5          & 53.0          & 55.0          & 66.5          & 51.9          & 47.6          & 54.0          \\
FWB~\cite{fwb}(ICCV'19)         &          & 47.0          & 59.6          & 52.6          & 48.3          & 51.9          & 50.9          & 62.9          & 56.5          & 50.1          & 55.1          \\ 
PFENet~\cite{pfenet}(TPAMI'20)         &          & 56.9          & 68.2         & 54.4          & \textbf{52.4}          & 58.0          & 59.0          & 69.1           & 54.8         & 52.9          & 59.0          \\ 
SCNet(ours)         &          & \textbf{58.0}          & \textbf{68.9}         & \textbf{57.0}          & 52.2          & \textbf{59.0}          & \textbf{59.8}          & \textbf{70.0}           & \textbf{62.7}         & \textbf{57.7}          & \textbf{62.6}          \\ \midrule
CANet~\cite{canet}(CVPR'19)       &          & 52.5          & 65.9          & 51.3          & 51.9          & 55.4          & 55.5          & 67.8          & 51.9          & 53.2          & 57.1          \\
PGNet~\cite{PGNet}(ICCV'19)       &          & 56.0          & 66.9          & 50.6          & 56.0          & 57.7          & 57.7          & 68.7          & 52.9          & 54.6          & 58.5          \\
CRNet~\cite{crnet}(CVPR'20)       &          & -             & -             & -             & -             & 55.7          & -             & -             & -             & -             & 58.8          \\
SimPropNet~\cite{simpropnet}(IJCAI'20) & ResNet-50 & 54.9          & 67.3          & 54.5          & 52.0            & 57.2          & 57.2          & 68.5          & 58.4          & 56.1          & 60.0           \\
RPMM~\cite{pmms}(ECCV'20)        &          & 55.2          & 66.9          & 52.6          & 50.7          & 56.3          & 56.3          & 67.3          & 54.5          & 51            & 57.3          \\
PFENet\cite{pfenet}(TPAMI'20)     &          & 61.7          & 69.5          & 55.4          & 56.3          & 60.8          & 63.1          & 70.7          & 55.8          & 57.9          & 61.9          \\
SCNet(ours)          &          & \textbf{62.2} & \textbf{70.5} & \textbf{61.1} & \textbf{58.1} & \textbf{63.0} & \textbf{63.3} & \textbf{72.0} & \textbf{68.4} & \textbf{60.2} & \textbf{66.0} \\ \midrule
FWB\cite{fwb}(ICCV'19)     &          & 51.3          & 64.5          & 56.7          & 52.2          & 56.2          & 54.8          & 67.4          & 62.2          & 55.3          & 59.9         \\ 
DAN(ECCV'20)     & ResNet-101         & 54.7          & 68.6          & 57.8          & 51.6          & 58.2          & 57.9          & 69.0          & 60.1          & 54.9          & 60.5         \\ 
PFENet\cite{pfenet}(TPAMI'20)     &          & 60.5          & 69.4          & 54.4          & 55.9          & 60.1          & 62.8          & 70.4          & 54.9          & 57.6          & 61.4          \\
SCNet(ours)     &          & \textbf{63.1}          & \textbf{71.1}          & \textbf{63.8}          & \textbf{57.9}          & \textbf{64.0}         & \textbf{67.5}          & \textbf{73.3}          & \textbf{67.9}          & \textbf{63.1}          & \textbf{68.0}          \\
\bottomrule
\end{tabular}}
\end{table*}

\begin{table*}[h]
\caption{Results of 1-shot and 5-shot segmentation on COCO-20$^i$ using the mean-IoU. Best results in bold.}
\label{coco}
\resizebox{\textwidth}{!}{
\begin{tabular}{@{}lccccccccccc@{}}
\toprule
                 &           & \multicolumn{5}{c|}{1-Shot}                                                 & \multicolumn{5}{c}{5-Shot}                            \\ \cmidrule(l){3-12} 
Methods           & Backbone  & Fold-0 & Fold-1 & Fold-2 & Fold-3 & \multicolumn{1}{c|}{Mean}          & Fold-0 & Fold-1 & Fold-2 & Fold-3 & Mean          \\ \midrule
FWB\cite{fwb}(ICCV'19)     &     & 18.4    & 16.7    & 19.6    & 25.4    & \multicolumn{1}{c|}{20.0}          & 20.9    & 19.2    & 21.9    & 28.4    & 22.6          \\ 
PANet\cite{panet}(ICCV'19)     & VGG-16    & -    & -    & -    & -    & \multicolumn{1}{c|}{20.9}          & -    & -    & -    & -    & 29.7          \\
PFENet\cite{pfenet}(TPAMI'20)     &     & 35.4    & 38.1    & 36.8    & 34.7    & \multicolumn{1}{c|}{36.3}          & 38.2    & 42.5    & 41.8    & 38.9    & 40.4          \\
\midrule
PPNet(ECCV'20)   &   & 34.5    & 25.4    & 24.3    & 18.6    & \multicolumn{1}{c|}{25.7}          & 48.3    & 30.9    & 35.7    & 30.2    & 36.2          \\
RPMM\cite{pmms}(ECCV'20)    & ResNet-50          & 29.5    & 36.8    & 29.0    & 27.0    & \multicolumn{1}{c|}{30.6}          & 33.8    & 42.0    & 33.0    & 33.3    & 35.5          \\ 
SCNet(ours)    &           & \textbf{35.7}    & \textbf{41.9}    & \textbf{37.2}    & \textbf{39.0}    & \multicolumn{1}{c|}{\textbf{38.4}}      & \textbf{39.6}    & \textbf{45.5}    & \textbf{41.9}    & \textbf{41.3}    & \textbf{42.1}          \\ 
\midrule
FWB\cite{fwb}(ICCV'19) &  & 19.9    & 18.0    & 21.0    & 28.9    & \multicolumn{1}{c|}{21.2}          & 19.1    & 21.5    & 23.9    & 30.1    & 23.7          \\
PFENet\cite{pfenet}(TPAMI'20) & ResNet-101 & 36.8    & 41.8    & 38.7    & 36.7    & \multicolumn{1}{c|}{38.5}          & 40.4    & 46.8    & 43.2    & 40.5    & 42.7          \\
SCNet(ours)      &           & \textbf{38.3}    & \textbf{43.1}    & \textbf{40.0}    & \textbf{39.1}    & \multicolumn{1}{c|}{\textbf{40.1}} & \textbf{44.0}    & \textbf{47.7}    & \textbf{45.0}    & \textbf{42.8}    & \textbf{44.8} \\ \bottomrule
\end{tabular}}
\end{table*}

\section{Experiments}  
\subsection{Experiment Setting}
\noindent\textbf{Datasets}
We evaluate the proposed approach on two public few-shot segmentation benchmarks: PASCAL-5$^i$~\cite{OSLSM} and COCO-20$^i$~\cite{fwb, panet}.
PASCAL-5$^i$ is built from PASCAL VOC 2012~\cite{voc} and extended annotations from SDS~\cite{sds}. This dataset contains 20 object classes divided into four folds and each fold has 5 categories. Following PFENet~\cite{pfenet}, 5000 support-query pairs were randomly sampled in each test fold for evaluation. We also evaluate our approach on a more challenging COCO-20$^i$, which is built on MS-COCO, as it contains more samples, more classes and more instances per image. Following~\cite{fwb}, COCO-20$^i$ also split four folds from 80 classes and each fold contains 20 categories. We use the same categories division and randomly sample 20,000 support-query pairs to evaluate as PFENet~\cite{pfenet}

For both datasets, we adopt 4-fold cross-validation that training model on three folds~(base class) and testing on remains one~(novel class). The experimental results are reported on each test fold. We also report the average performance of all four test folds.

\noindent\textbf{Evaluation Metric}
We use the widely adopted the mean intersection over union~($\mathrm {mIoU}$) for quantitative evaluation. For each class, the $\mathrm {IoU}$ is calculated by $\frac{\mathrm{TP}}{\mathrm{TP}+\mathrm{FP}+\mathrm{FN}}$, where $\mathrm {TP}$ is the number of true positives, $\mathrm{FP}$ is the number of false positives and $\mathrm{FN}$ is the number of false positives over the prediction and ground-truth masks on query set. The $\mathrm {mIoU}$ is an average of all different classes $\mathrm{IoU}$, \ie, $\mathrm{mIoU}$=$\frac{1}{n_{c}} \sum_{i} \mathrm{IoU}_{i}$, where ${n_{c}}$ is the number of novel classes.

\subsection{Implementation Details}
Our approach builds on PFENet~\cite{pfenet} with ResNet-50 and ResNet-101 as backbones for a fair comparison with other
methods. These backbone networks are initialized with ImageNet pre-trained weights and we keep their weights fixed during training. Other layers are initialized by the default setting of PyTorch. The network is trained on PASCAL-5$^i$ with the initial learning of 0.0025 and the momentum of 0.9 for 200 epochs with 4 pairs of support-query images per batch. For COCO-20$^i$, model are trained for 50 epochs with a learning rate of 0.005 and batch size 8. We randomly crop 473$\times$473 patches from the processed images as training samples. The $k$-means algorithm iterates 10 rounds to calculate the pseudo mask for query image. Data augmentation strategies including normalization, mirror operation and random rotation from -10 to 10 degrees are used. Our experiments do not use any post-processing techniques to refine the results. All experiments are conducted on NVIDIA Tesla V100 GPUs and Intel Xeon CPU Platinum 8255C.

\begin{figure*}[t]
\centering
\centerline{\includegraphics[width=1\linewidth]{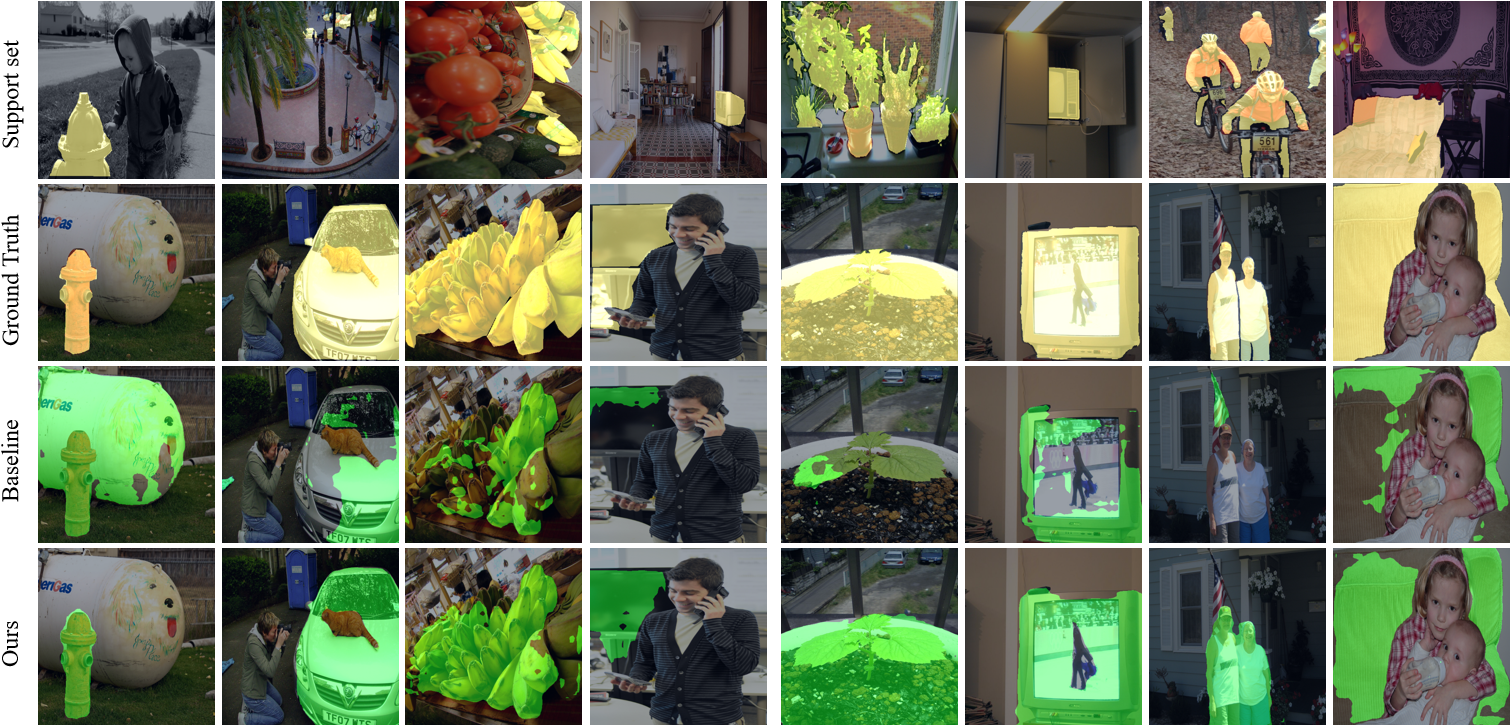}}
\caption{Qualitative results of the proposed SCNet and the baseline. The left samples are from COCO-20$^i$ and the right ones are from PASCAL-5$^i$. The first and second row is support and query images with their ground-truth annotations. The third and fourth row is segmentation results of baseline and our SCNet.}
\label{fig:qvis}
\end{figure*}

\subsection{Comparison with State-of-the-Arts} 
As reported in Tables~\ref{table:pascal} and~\ref{coco},
we compare the
proposed method with state-of-the-arts on PASCAL-5$^i$ and
COCO-20$^i$. We can see that our
method significantly outperforms state-of-the-art approaches in both 1-shot and 5-shot settings. Additional qualitative results are shown in Fig.~\ref{fig:qvis}.

\noindent\textbf{PASCAL-5$^i$ Results.} 
We report the $\mathrm {mIoU}$ of each fold and the mean of all four folds on PASCAL-5$^i$ in Table~\ref{table:pascal}. We can see that the SCNet significantly outperforms state-of-the-arts with all backbones. Using ResNet-50, ours method achieves 2.2\% (63.0\% vs.60.8\%) and 4.1\% (66\% vs.61.9\%) improvements compared with PFENet in the 1-shot and 5-shot settings, respectively. This indicates that exploiting complete feature comparison is clearly beneficial for few-shot semantic segmentation.

\noindent\textbf{COCO-20$^i$ Results.}
The results on COCO-20$^i$ are reported in Table~\ref{coco}. Our method performs competitively with state-of-the-art approaches in the 1-shot setting and significantly outperforms recent methods in the 5-shot scenario.  These results suggest that our additional class-agnostic branch constrains the model to learn better feature representation and yield an unbiased classifier, while other methods may tend to predict some novel class objects as the background even if more support information is supplied.

\noindent\textbf{Qualitative Results.} 
We show some qualitative results on the PASCAL-5$^i$ and COCO-20$^i$ test set in Fig.~\ref{fig:qvis}. First, our method is capable of making correct predictions even if the background of the query image contains some other targets, \eg, the person in the second and fourth images. Second, the proposed method has a better generalization performance when there is diverse semantics among objects in the support-query images. Note that the baseline method may segment parts of the novel class object because of the prior bias, that hinders the segmentation of the novel class object. Third, our method may fail when some small objects are presented in the foreground, \eg, the person on TV.
\subsection{Ablation Studies}  
In order to comprehend how SCNet works, we perform exhaustive experiments to analyze the components in SCNet. All experiment results are evaluated over all folds of the PASCAL-5$^i$ using the ResNet-50 backbone.

\noindent\textbf{Number of Background Prototype.} 
We fix the loss weight $\lambda$ to 0.5 and choose the value
$n$ (the number of clusters) from a given set $\{1, 2, 3, 4, 5\}$. As shown in Table~\ref{Number of $n$}, first, our method performs better than the baseline under all cases even in the simplest case~(\ie, $n=1$). Second, the best performance is achieved when $n$ is set to 3, which validates the introduction of clustering to obtain more fine prototypes. Third, the performance slightly decreases when $n$ is either too small or too large. We argue that: when $n$ is too small, the feature comparison becomes difficult, as background regions may contain some different semantic information; and when $n$ is too large, the learning task will be so simple and easily fall into over-fitting, due to limited semantics on the query images.
\begin{table}[t]
\centering
\caption{Performance~(mIoU\%) comparison with different values of $n$~(the number of background prototype). $n$ = 0 is equivalent to the baseline method.}
\label{Number of $n$}
\begin{tabular}{@{}c|cccc|c@{}}
\toprule
$n$      & \multicolumn{1}{l}{Fold-0} & \multicolumn{1}{l}{Fold-1} & \multicolumn{1}{l}{Fold-2} & \multicolumn{1}{l|}{Fold-3} & \multicolumn{1}{l}{Mean} \\ \midrule
0      & 61.7                      & 69.5                      & 55.4                      & 56.3                       & 60.8                    \\ 
1      & 61.7                      & 70.1                      & 60.4             & 55.0                       & 61.8                    \\
2      & 61.9                      & 70.0                      & 61.6                      & 55.8                       & 62.3                    \\
3      & \textbf{62.2}             & \textbf{70.5}             & 61.1                      & \textbf{58.1}              & \textbf{63.0}           \\
4      & 62.0                      & 70.3                      & \textbf{62.0}                      & 56.1                       & 62.6                    \\
5      & 61.3                      & 70.1                      & 60.1                      & 56.6                       & 62.0     \\ \bottomrule
\end{tabular}
\end{table}

\begin{table}[h]
\centering
\caption{Effect of background prototype generation method.}
\label{table:source}
\begin{tabular}{@{}c|ccccc@{}}
\toprule
Prototype & Fold-0 & Fold-1 & Fold-2 & Fold-3 & Mean \\ \midrule
Support  & 59.8    & 69.7    & 58.2    & \textbf{56.7}    & 61.1 \\
Query  & \textbf{61.7}        &  \textbf{70.1}       &  \textbf{60.4}       & 55.0        & \textbf{61.8}     \\ \bottomrule
\end{tabular}
\end{table}

\begin{table}[t]
\caption{The influences of hyper-parameter ($\lambda$) for our SCNet.}
\label{table:loss-weight}
\centering
\begin{tabular}{@{}c|cccc|c@{}}
\toprule
$\lambda$ & Fold-0         & Fold-1         & Fold-2         & Fold-3         & Mean           \\ \midrule
0      & 61.7          & 69.5          & 55.4          & 56.3          & 60.8          \\
0.1    & 61.1          & 69.9          & 58.2          & 57.3          & 61.6          \\
0.3    & 60.8          & 70.1          & 60.5          & 57.0          & 62.1          \\
0.5    & \textbf{62.2} & \textbf{70.5} & 61.1          & \textbf{58.1} & \textbf{63.0} \\
0.7    & 61.2          & 70.3          & \textbf{61.4} & 57.2          & 62.5          \\
0.9    & 59.8          & 69.2         & 60.0          & 53.3          & 57.7          \\
1.0    & 17.2          & 28.8          & 33.2          & 21.0          & 25.1          \\ \bottomrule
\end{tabular}
\end{table}

\noindent\textbf{Effects of Loss Weight.}
We explore the influence of hyper-parameter $\lambda$, where $\lambda$ is a weight that balances the importance between
two losses. From Table~\ref{table:loss-weight}, 
first, we can see that, when $\lambda$ is 0.5, \ie the two losses are equally important, our method yields a 2.2 mIoU improvement over the baseline. This implies the effectiveness of the second loss $L_2$. Second, the performance begins to decrease when $L_2$ gets more attention. Extremely, when we set $\lambda$ to 1, the performance drops rapidly (from 63.0 to 25.1). The reason is that now the whole network learns feature comparison only from the query image without any information from the support images. Third, the performance drops as $\lambda$ decreases. When $\lambda$ is 0, the overall network will degenerate to the baseline which learns feature comparison only on support-query image pair. 

\noindent\textbf{Background Prototype Selection.}
\begin{table}[t]
\centering
\caption{Results using spatial pyramid pooling prototypes.}
\label{table:spp}
\begin{tabular}{@{}c|cccc|c@{}}
\toprule
Spatial Size    & Fold-0 & Fold-1 & Fold-2 & Fold-3 & Mean \\ \midrule
2$\times$2 & 60.1    & 69.1    & 57.4    & 54.0    & 60.1 \\
3$\times$3 & 60.6    & 69.2    & 60.2    & 55.6    & 61.4 \\
4$\times$4 & \textbf{60.9}    & \textbf{70.0}    & \textbf{60.7}    & 56.1    & \textbf{61.9} \\
5$\times$5 & 59.8        &  69.7       &  58.2       & \textbf{56.7}        & 61.1     \\ \bottomrule
\end{tabular}
\end{table}
Instead of extracting background prototypes from support features~\cite{panet, PL, fwb, simpropnet}, we generate background prototypes from the query features to completely learn feature comparison. Which one is better? To answer this problem, we conduct experiments base on support/query background prototypes. The number of clusters is set to 1 in our SCNet to eliminate the influence of clustering. Table~\ref{table:source} reports the results of these two settings. We can see that using the background prototype from query feature outperforms the counterpart. This is because it is not appropriate to derive from the support image the background prototype for the query image as we discussed. 

\noindent\textbf{Importance of Prototype Generation Method.}
In our method, the background prototypes are generated by using the $k$-means clustering algorithm on the query feature map. We conduct another generation approach to see whether the performance is sensitive to the generation strategy. Different from the clustering algorithm which adapts the assignments according to spatial semantic distribution, we simply apply spatial pyramid pooling~\cite{he2015spatial} on a query feature map to obtain background prototypes. With a pyramid level of a$\times$a bins, we implement the average pooling at each bin and obtain a total of $a^2$ prototypes. In our experiment, we use 4-level pyramid: \{2$\times$2, 3$\times$3, 4$\times$4, 5$\times$5\}. 

In Table~\ref{table:spp}, we can see that using the spatial background prototypes generation is comparable to the baseline (61.9 vs.~60.8). This further demonstrates the effectiveness of learning a complete feature comparison. Besides, our method is still higher than spatial assignment under all spatial sizes (63.0 vs.~61.9), which implies that it is better to use a semantic clustering strategy for background prototype generation.

\section{Conclusion}  

In this paper, we embed complementary feature comparison into metric-based few-shot semantic segmentation (FSS) framework to improve the FSS performance.
Specifically, unlike previous works unilaterally predicting foreground mask with the prototypes extracted for foreground objects only, we propose to compute background prototypes and construct complementary sample pairs, which enables us to perform a complementary feature comparison with two-branch network architecture, \ie the class-specific and class-agnostic branch. To ensure the similarity between the prototype and the background features during prediction,
we propose to extract background prototypes from the query features alone and feature comparison is conducted in a self-contrastive manner.
The proposed network SCNet achieves state-of-the-art performance on both PASCAL-5$^i$ and COCO-20$^i$ datasets, which validates the effectiveness of our method.
{\small
\bibliographystyle{ieee_fullname}
\bibliography{CCFNet.bbl}
}

\end{document}